\documentclass[10pt,twocolumn,letterpaper]{article}

\usepackage[pagenumbers]{wacv} 

\definecolor{wacvblue}{rgb}{0.21,0.49,0.74}
\usepackage[pagebackref,breaklinks,colorlinks,allcolors=wacvblue]{hyperref}
\usepackage{multirow}
\usepackage{amssymb}
\usepackage{tabularray}
\UseTblrLibrary{booktabs}
\usepackage{rotating}
\usepackage[table]{xcolor} 
\usepackage{booktabs} 
\def\wacvPaperID{723} 
\def\confName{WACV}
\def\confYear{2026}

\title{MedROV: Towards Real-Time Open-Vocabulary Detection Across Diverse Medical Imaging Modalities}

\author{Tooba Tehreem Sheikh
\and Jean Lahoud \and Rao Muhammad Anwer \and Fahad Shahbaz Khan \and Salman Khan  \and Hisham Cholakkal \vspace{1pt}  \and 
{Mohamed Bin Zayed University of Artificial Intelligence (MBZUAI) } \and
{\tt \small \{tooba.sheikh, jean.lahoud, rao.anwer, fahad.khan, salman.khan, hisham.cholakkal\}} \and \tt \small @mbzuai.ac.ae
}

\begin{document}
\maketitle
\begin{abstract}
Traditional object detection models in medical imaging operate within a closed-set paradigm, limiting their ability to detect objects of novel labels. Open-vocabulary object detection (OVOD) addresses this limitation but remains underexplored in medical imaging due to dataset scarcity and weak text-image alignment. To bridge this gap, we introduce MedROV, the first Real-time Open Vocabulary detection model for medical imaging. To enable open-vocabulary learning, we curate a large-scale dataset, Omnis, with 600K detection samples across nine imaging modalities and introduce a pseudo-labeling strategy to handle missing annotations from multi-source datasets. Additionally, we enhance generalization by incorporating knowledge from a large pre-trained foundation model. By leveraging contrastive learning and cross-modal representations, MedROV effectively detects both known and novel structures. 
Experimental results demonstrate that MedROV outperforms the previous state-of-the-art foundation model for medical image detection with an average absolute improvement of 40 mAP50, and surpasses closed-set detectors by more than 3 mAP50, while running at 70 FPS, setting a new benchmark in medical detection. Our source code, dataset, and trained model are available at \href{https://github.com/toobatehreem/MedROV}{MedROV}. 
\end{abstract}
    
\begin{figure}[t]
    \centering
    \includegraphics[width=\linewidth]{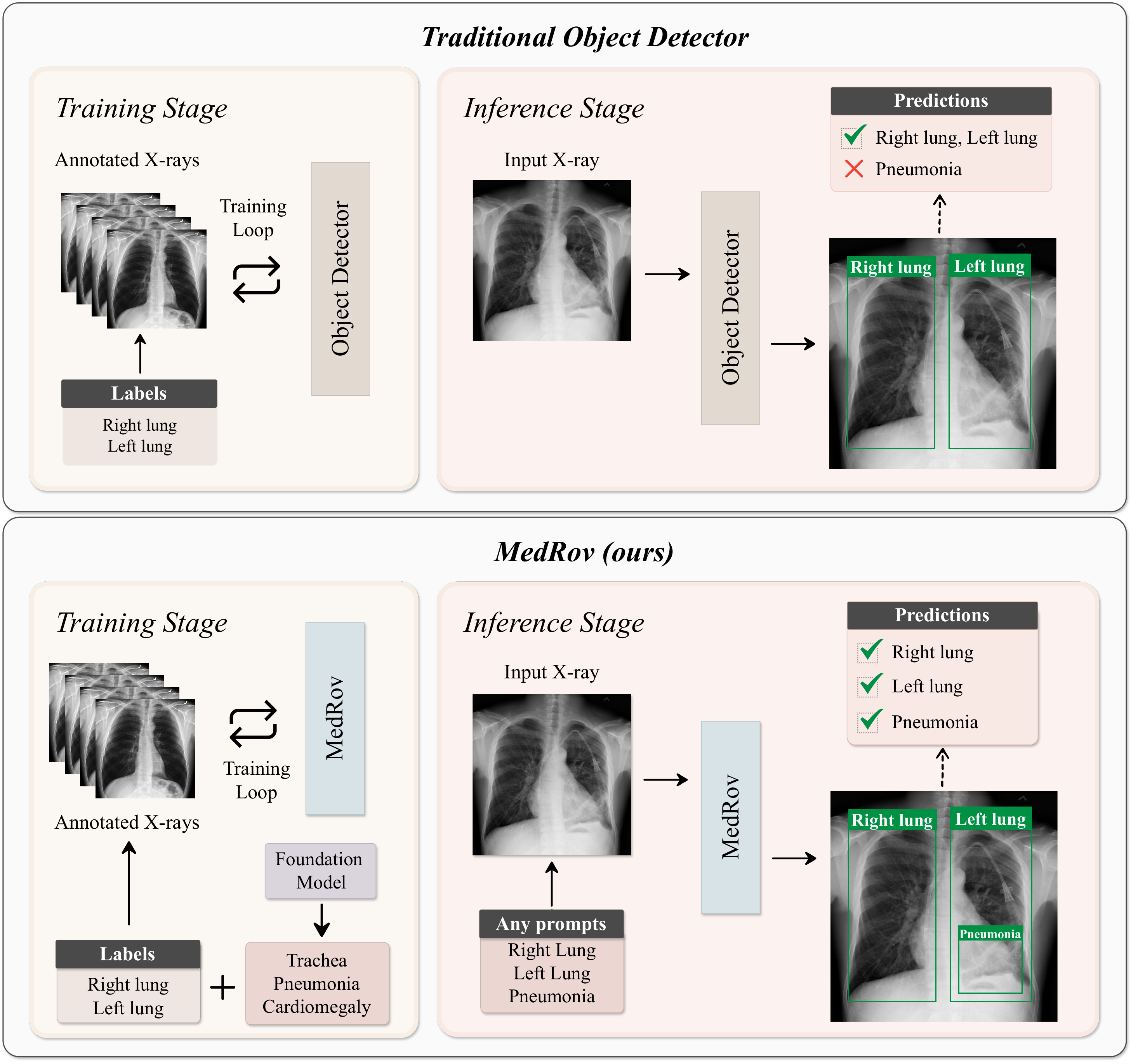}
    \caption{\textbf{Comparison of Traditional Object Detection and MedROV:} Traditional detectors are trained on a fixed set of categories and cannot recognize unseen classes. For example, the model shown here detects the left and right lungs but fails to detect pneumonia, which is present in the image. In contrast, MedROV is a Real-time Open Vocabulary detection model for medical imaging that leverages the BioMedCLIP foundation model to enable detection of both seen and unseen classes. At inference, it can detect any class described by a text prompt, if present in the image.}
    \label{fig:ovod}
\end{figure}

\section{Introduction}
\label{sec:intro}

Object detection in medical imaging plays an important role in identifying abnormalities such as tumors, fractures, and diseased cells across diverse modalities, including CT scans, X-rays, MRIs, and histopathology slides.
Unlike natural images, medical imaging poses unique challenges due to its multi-modal nature, where each modality has distinct visual and semantic characteristics. 

Traditional object detection models, including single-stage \cite{redmon2016you,zhao2024detrs} and two-stage \cite{ren2016faster,he2017mask} detectors, have been adapted for medical tasks like tumor detection and organ localization. However, these models are limited to detecting only predefined categories (closed-set detection), making them ineffective in real-world medical scenarios where new, critical abnormalities may emerge and require immediate detection. 
On the other hand, Open Vocabulary Object Detection (OVOD) addresses the limitations of closed-set detection models by enabling the detection of novel objects through vision-language alignment \cite{cheng2024yolo} or region-level vision-language pre-training \cite{liu2024grounding}. While OVOD has shown significant potential in natural images, its application to medical imaging remains largely unexplored. Adapting OVOD methods, such as YOLO-World, to the medical domain presents challenges due to the scarcity of large, diverse, and well-annotated image-text datasets necessary for learning meaningful visual-language associations. Additionally, medical images exhibit complex variations in object size and shape, overlapping objects, and imbalances in class distribution, making the application of OVOD in the medical field more challenging.

To address these challenges, we introduce MedROV, the first Real-time Open Vocabulary detection model for medical imaging, designed to detect both known and novel structures across nine imaging modalities (Fig. \ref{fig:ovod}). Trained on a large-scale dataset of over 600K samples, our model outperforms existing OVOD methods, previous state-of-the-art medical detection methods, and closed-set detectors by leveraging the BioMedCLIP \cite{zhang2023biomedclip} foundation model to enhance open vocabulary detection. In summary, our main contributions are as follows: 
\begin{itemize}
    \item We introduce the first open-vocabulary object detector for medical images, capable of detecting both known and unknown structures, by curating Omnis, a large-scale dataset of over 600K detection samples across 9 imaging modalities (CT, MRI, X-ray, Ultrasound, Histopathology, Dermoscopy, Fundoscopy, Endoscopy, and Microscopy).
    \item We adapt YOLO-World, an open vocabulary detector originally designed for natural images, to the medical domain by training it on our dataset and addressing the challenge of missing annotations when integrating multiple datasets across different modalities.
    \item We improve the model's detection and generalization performance by incorporating information from the BioMedCLIP foundation model, leveraging its vision-language features for improved performance.
    \item We conduct extensive experiments comparing our method with existing OVOD approaches, the previous state-of-the-art model \cite{zhao2024biomedparse}, and closed-set detectors. Our model achieves significant improvement in zero-shot detection performance on the Omnis test set compared to the baseline YOLO-World, while maintaining comparable speed (YOLO-World: 72 FPS, Ours: 70 FPS). Additionally, it outperforms the BioMedParse foundation model by an average absolute improvement of 40 mAP50 and surpasses closed-set methods by more than 3 mAP50.
\end{itemize}
\section{Related Work}
\label{sec:relatedwork}
Recent advances in deep learning have significantly improved performance across various medical imaging tasks, including classification, segmentation, and detection. Object detection, in particular, plays a crucial role in identifying anatomical structures and pathological abnormalities within medical scans. While traditional object detectors have been successfully adapted from natural images to medical domains, they largely operate under closed-set assumptions, limiting their ability to generalize to unseen or rare categories. Meanwhile, the emergence of large-scale foundation models and open-vocabulary object detection techniques has enabled the development of more flexible and generalizable systems. In this section, we review prior work in three relevant areas: object detection in medical imaging, the application of foundation models to medical tasks, and recent progress in open-vocabulary object detection.
\begin{figure*}[!ht]
    \centering
     \includegraphics[width=\linewidth, trim={0 2mm 0 3mm},clip]{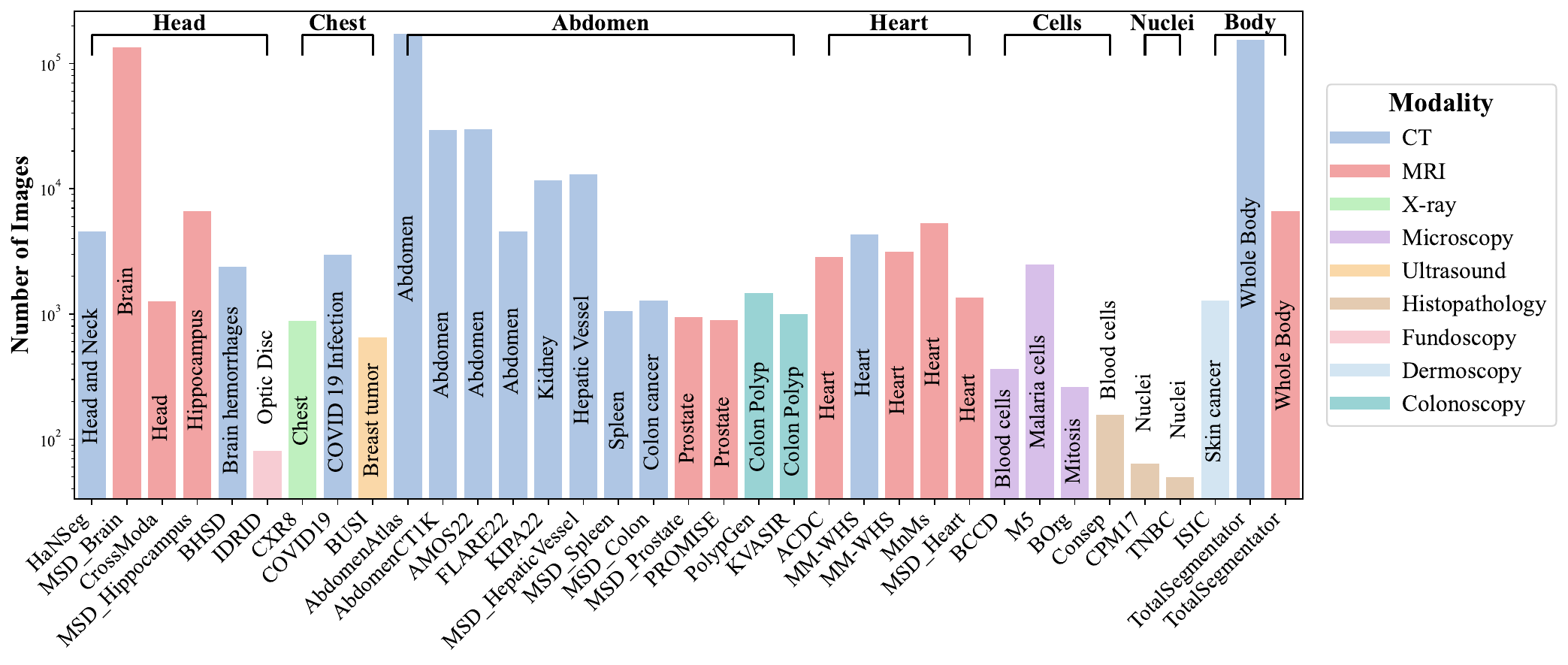}
    \caption{\textbf{An overview of the Omnis 600K dataset.}
    We curate a large-scale object detection dataset for medical imaging by incorporating 35 datasets with diverse modalities (represented in different colors), anatomical regions (displayed at the top), and target areas (indicated within the bars).}
    \label{fig:datasetfig}
\end{figure*}
\subsection{Object Detection in Medical Images}
Object detection is essential in medical imaging, allowing for the accurate detection of tumors, lesions, and other abnormalities. Deep learning-based detectors in natural images have been adapted for medical imaging. RT-DETR with multi-scale feature extraction has been applied to diabetic retinopathy detection \cite{he2025object}, while BGF-YOLO \cite{kang2024bgf} and SOCR-YOLO \cite{liu2024socr} have introduced enhancements to the YOLO architecture for brain tumor detection and lesion detection, respectively. 
To further enhance detection accuracy across varying object sizes and complex backgrounds, recent work by \cite{xu2025cross} proposed a cross-scale attention and multi-layer feature fusion method based on YOLOv8 for skin disease detection.
However, these approaches remain limited to closed-set object detection, restricting their ability to identify novel or previously unseen structures.

\subsection{Foundation Models in Medical Images}
Recent foundation models have advanced medical imaging through multi-modal learning and self-supervised training. MedSAM \cite{ma2024segment} adapts SAM for medical image segmentation, while BioMedCLIP \cite{zhang2023biomedclip}, trained on the PMC15M dataset, enhances CLIP for biomedical image understanding. MEDITRON \cite{chen2023meditron}, a large-scale medical language model, outperformed previous state-of-the-art models on USMLE-style questions. 
MedPaLM-2 \cite{singhal2025toward}, an adaptation of PaLM-2 for healthcare, demonstrates strong performance in clinical question answering. Additionally, BioMedParse \cite{zhao2024biomedparse}, a biomedical foundation model for image parsing, introduces a unified approach for joint segmentation, detection, and recognition. However, most of these models remain limited to classification and question answering tasks, restricting their ability to tackle more complex and open-ended challenges in medical imaging.

\subsection{Open-Vocabulary Object Detection} 
Open-Vocabulary Object Detection (OVOD) extends traditional detection by identifying both known and novel objects through semantic understanding. Significant progress has been made on natural images with models such as DINO-X \cite{ren2024dino}, which enhances open-world detection using a Transformer-based encoder-decoder architecture, and DetCLIPv3 \cite{yao2024detclipv3}, which integrates open-set detection with captioning for detailed descriptions. Detic \cite{zhou2022detecting} leverages image-level labels and large vocabularies to perform open-vocabulary detection and classification. GLIP \cite{li2022grounded} enables zero-shot generalization by formulating detection as a matching task between image regions and text. Additionally, YOLO-World \cite{cheng2024yolo} integrates the lightweight YOLO framework with CLIP for real-time detection. Despite these advancements, OVOD in medical imaging remains largely unexplored. Our approach builds upon YOLO-World by incorporating the BioMedCLIP foundation model to enhance vision-language alignment and improve OVOD performance in medical imaging.

\section{Methodology}
\label{sec:methodology}
\begin{figure*}[t]
    \centering
    \includegraphics[width=\linewidth, trim={5mm 2 5mm 2},clip]{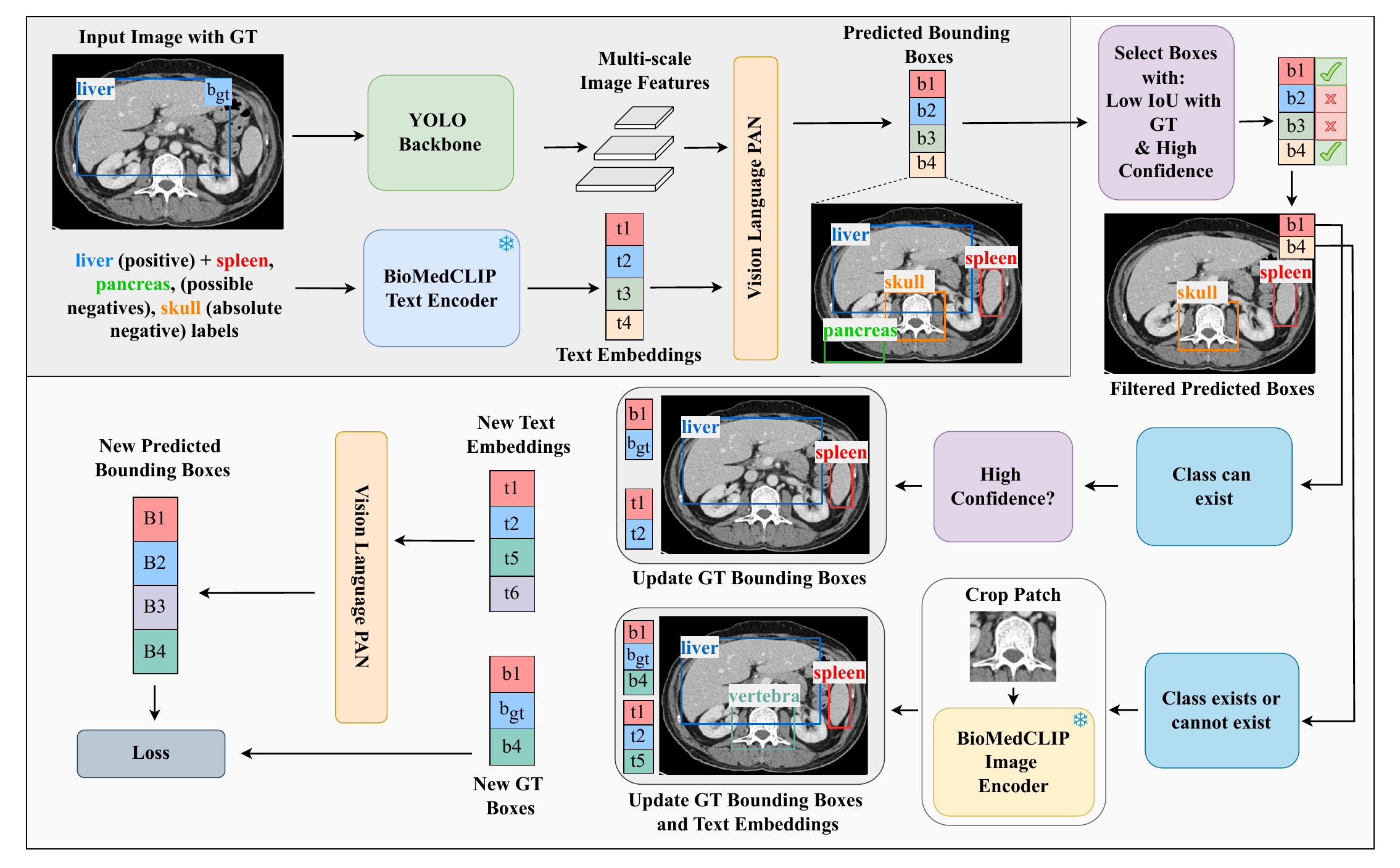}
 \caption{\textbf{Overall architecture of MedROV.}
 The model takes image and text labels as input. During training, positive and negative labels are used, whereas during testing, free-form text labels can be passed. 
 The YOLO backbone extracts image features, while the BioMedCLIP text encoder generates text embeddings. 
 These features are fused using the Vision-Language PAN (VL-PAN) to obtain bounding box predictions. 
 During training (outer box), predictions are first filtered based on an IoU threshold. If a predicted class is missing in the dataset but can exist, the high-confidence bounding box is added to the ground truth as pseudo-label. 
 Otherwise, the cropped region is passed through the BioMedCLIP image encoder for feature extraction. The extracted features replace one of the negative text label embeddings, updating the ground truth. The updated text embeddings and bounding boxes are passed through VL-PAN again to generate refined predictions. Finally, the loss is computed between the new ground truth and the updated predictions. 
 }
\label{fig:medyoloworld-arch}
\end{figure*}

We introduce MedROV, a Real-time Open Vocabulary detection model for medical imaging. In our work, we build upon the baseline YOLO-World \cite{cheng2024yolo} (Section \ref{sec:baseline}) and incorporate domain-specific adaptations.
To this end, we curate a large-scale dataset of 600K detection samples spanning nine imaging modalities, detailed in Section \ref{sec:dataset}.
We address the missing annotation challenge through pseudo-labeling, and make use of a large-scale medical foundation model, BioMedCLIP \cite{zhang2023biomedclip}, to improve generalizability (Section \ref{sec:medyoloworld}).
\subsection{Baseline} \label{sec:baseline}
YOLO-World is an open-vocabulary object detection framework that extends the YOLOv8 architecture \cite{redmon2016you} by integrating vision-language modeling to enable detection beyond a fixed set of categories. It incorporates a CLIP-based text encoder to extract text embeddings from user-defined prompts and a Re-parameterizable Vision-Language Path Aggregation Network (RepVL-PAN) for cross-modal fusion of text and image features.

During training, YOLO-World constructs region-text pairs by assigning both positive and negative text labels to image regions. Positive labels correspond to objects present in the image, while negatives are randomly sampled from the remaining dataset classes that are not in the image, forming a dynamic vocabulary of size $M$ labels per image. A region-text contrastive loss $\mathcal{L}_{\mathrm{Con}}$ is employed to optimize alignment between predicted object embeddings and text embeddings, where the object-text similarity is computed via an L2-normalized dot product followed by an affine transformation: $s_{k,j} = \alpha \cdot \text{L2-Norm}(e_k) \cdot \text{L2-Norm}(w_j)^\top + \beta$, where \(e_k\) is the predicted object embedding, \(w_j\) is the text embedding, and \(\alpha, \beta\) are learnable scaling and shifting parameters. The model also incorporates an IoU loss (\(\mathcal{L}_{\text{IoU}}\)) and a Distributed Focal Loss (\(\mathcal{L}_{\text{DFL}}\)) for bounding box regression. The total loss is given by:
\(\mathcal{L}(I) = \mathcal{L}_{\text{Con}} + \lambda_I \cdot (\mathcal{L}_{\text{IoU}} + \mathcal{L}_{\text{DFL}})
\), where \(\lambda_I\) is an indicator set to 1 for samples with reliable box annotations (e.g., detection or grounding data) and 0 for weakly labeled image-text data. While YOLO-World excels in natural image detection, its performance on medical images is limited due to the domain gap, multi-modality nature of medical imaging, and a lack of large annotated medical datasets.

\subsection{Omnis Dataset}
\label{sec:dataset}
Medical imaging spans multiple modalities with significant domain gaps. To enable open-vocabulary detection in MedROV, we curate Omnis 600K, a large-scale medical detection dataset covering nine imaging modalities and diverse anatomical and pathological targets (Fig. \ref{fig:datasetfig}). It integrates public datasets from CT \cite{qu2024abdomenatlas,ma2021abdomenct,ji2022amos,wu2023bhsd,ma2021toward,ma2022fast,podobnik2024han,zhuang2018multivariate,antonelli2022medical,wasserthal2023totalsegmentator}, MRI \cite{bernard2018deep,dorent2023crossmoda,campello2020multi}, Ultrasound \cite{al2020dataset}, Histopathology \cite{graham2019hover,naylor2018segmentation}, X-ray \cite{wang2017chestx}, Colonoscopy \cite{pogorelov2017kvasir,liu2025polyp}, Microscopy \cite{BCCD,awais2024borg,sultani2022towards}, Dermoscopy \cite{codella2019skin}, and Fundoscopy \cite{porwal2020idrid}.

For 3D modalities like CT and MRI, each volume is processed into in-plane 2D slices for consistency. Segmentation masks are converted to detection bounding boxes by identifying the minimum and maximum coordinates of non-zero regions per slice. To handle varying intensity distributions, we apply modality-specific normalization. Following MedSAM \cite{ma2024segment}, we clip CT intensities to [-500, 1000] and MRI intensities to the 0.5th–99.5th percentile range. All images are normalized to a [0, 255] scale. Grayscale images are converted to 3-channel by replicating the single channel.

Omnis is specifically designed for open-vocabulary object detection. It comprises 157 training classes and incorporates the BioMedCLIP foundation model to enable generalization beyond fixed categories. Omnis consists of 577k training and 28k validation images. Following MedSAM \cite{ma2024segment} and BioMedParse \cite{zhao2024biomedparse}, we split the data at the volume level to prevent data leakage, ensuring slices from the same 3D scan remain within a single split. We allocate 95\% of the data for training and 5\% for validation. We also hold out some classes within Omnis, ensuring that entire volumes containing these classes are excluded from training for zero-shot evaluation. This held-out set is included in our test set (see Section \ref{sec:experimentsandresults} for details).
Unlike datasets such as BioMedParse \cite{zhao2024biomedparse}, which is limited to 82 predefined categories and follows an image–mask–label triplet format, constraining its ability to generalize to broader tasks, or MedSAM, which lacks detection capabilities and a scalable vocabulary, Omnis is specifically designed for medical OVOD.

\subsection{MedROV Architecture} \label{sec:medyoloworld}
\textbf{Overview:} MedROV, illustrated in Fig. \ref{fig:medyoloworld-arch}, leverages BioMedCLIP \cite{zhang2023biomedclip} as its text encoder, as it is specifically designed for medical applications and trained on a large-scale dataset of medical image–text pairs, making it well-suited for medical image understanding. Our method introduces open-vocabulary capability and enhances performance by training the model on the large-scale Omnis 600K detection dataset and addressing the missing annotations problem in medical datasets through pseudo-labeling. Additionally, we integrate BioMedCLIP features into MedROV for feature alignment (detailed next). \\
\noindent\textbf{Addressing Missing Annotations:} 
Medical datasets are typically labeled for specific targets, resulting in many object labels being present in images but not annotated. When combining multiple datasets, the number of labels increases, leading to more missing annotations. The absence of these annotations would lead to penalizing the model during training for predicting visible but unannotated objects, causing confusion and reducing learning efficiency. 
To address this issue, we introduce a dataset-class presence matrix (denoted by $M$), which categorizes the availability of annotations for each class across datasets.
The matrix \( M \) assigns a value to each dataset-class pair \( (d, c) \) as follows: \( M_{d,c} = 1 \) if class \( c \) is annotated in dataset \( d \), \( M_{d,c} = 0 \) if class \( c \) may be present in \( d \) but is not annotated, and \( M_{d,c} = -1 \) if class \( c \) cannot exist in \( d \).

During training, we extract predictions from the detection head and apply non-maximum suppression (NMS) to eliminate redundant boxes. To identify unannotated objects, we compute the Intersection over Union (IoU) between each predicted bounding box and all ground truth boxes, regardless of the class labels (class-agnostic). 
A prediction is considered a potential missed annotation if its maximum IoU with any ground truth box falls below a predefined threshold $T$ i.e.: $\max_{g \in G} \text{IoU}(p, g) < T$, where $p$ is the predicted bounding box, and $G$ is the set of ground truth boxes. This indicates that the model has detected an object that does not overlap with any annotated ground truth. 

For each potentially missed annotation, we check $ M_{\text{d}, \text{c}} $. If $M_{\text{d}, \text{c}}$ = 0, indicating that class $\text{c}$ may be present but is not annotated in dataset $\text{d}$, and the model's prediction confidence exceeds a threshold $C$, the prediction is added to the ground truth as a pseudo-label to prevent the model from being incorrectly penalized for detecting valid but unannotated objects, improving training stability and performance.

\noindent\textbf{Enhancing Generalization Using Foundation Medical Image Models:}
To enhance the generalizability and open-vocabulary capability of MedROV, we incorporate knowledge from BioMedCLIP \cite{zhang2023biomedclip}, a foundation model trained on a large-scale medical dataset of image–text pairs.
Given a prediction, we refer to the dataset-class matrix $M_{\text{d}, \text{c}}$. If $M_{\text{d}, \text{c}} = 1$ (class annotated) or $M_{\text{d}, \text{c}} = -1$ (class cannot exist), we refine predictions by discarding overly small boxes, full-image boxes, or those that predominantly cover background regions (e.g., boxes containing mostly black pixels).
The remaining boxes are expanded by a factor $F$ = 1.3, and the cropped region is processed through the BioMedCLIP image encoder to align image features with textual representations.
For example, if the model predicts a liver in a dataset where liver annotations already exist ($M_{\text{d}, \text{c}} = 1$), but the actual object is, say, a spleen, BioMedCLIP helps ensure the extracted visual features align more closely with the spleen’s textual features. Similarly, if the model predicts a skull in an abdomen dataset, where such an object cannot exist ($M_{\text{d}, \text{c}} = -1$), BioMedCLIP encourages alignment with the correct object’s text features, correcting the semantic mismatch.
To avoid direct reliance on noisy or incorrect labels, we replace the text features of a negative ground truth sample with the extracted BioMedCLIP image features and add the bounding box to the ground truth, repeating this for up to five boxes per image, sorted by confidence. By training the model on semantically grounded features instead of potentially incorrect labels, this approach mitigates error propagation, enhances robustness, and improves the open-vocab capability and generalizability of MedROV.

The final loss, identical to that of the baseline YOLO-World \cite{cheng2024yolo}, is computed between the updated ground truth and the model predictions. During inference, MedROV adopts the prompt-then-detect paradigm of YOLO-World \cite{cheng2024yolo}, enabling real-time detection with comparable speed across diverse medical imaging tasks.
\begin{table*}[t]
\centering \small
\resizebox{1\textwidth}{!}
{
\begin{tabular}{ l | l | l | c c | c c | c c }
\toprule
\rowcolor[gray]{0.85}
\textbf{Models} $\rightarrow$ & & & \multicolumn{2}{c|}{\textbf{YOLO-World}} & \multicolumn{2}{c|}{\textbf{YOLO-World + Our Omnis}} & \multicolumn{2}{c}{\textbf{MedROV (Ours)}} \\
\rowcolor[gray]{0.85}
\textbf{Dataset} $\downarrow$ & \textbf{Modality} & \textbf{Classes} & \textbf{mAP50} & \textbf{mAP50:95} & \textbf{mAP50} & \textbf{mAP50:95} & \textbf{mAP50} & \textbf{mAP50:95} \\
\midrule

\multirow{2}{*}{BTCV}             & \multirow{2}{*}{CT} & Base & 0.00 & 0.00 & 80.7 & 62.2 & \textbf{83.4} & \textbf{63.1} \\
                                  &                     & Base + Novel & 0.03 & 0.01 & 74.5 & 57.5 & \textbf{79.1} & \textbf{59.3} \\
\midrule

\multirow{2}{*}{Cervix}           & \multirow{2}{*}{CT} & Base & 0.00 & 0.00 & 64.3 & 39.3 & \textbf{66.9} & \textbf{42.3} \\
                                  &                     & Base + Novel & 0.00 & 0.00 & 33.3 & 20.7 & \textbf{33.8} & \textbf{21.3} \\
\midrule
\multirow{2}{*}{MSD Liver} & \multirow{2}{*}{CT} & Base         & 0.34 & 0.09 & \textbf{99.4} & \textbf{94.7} & \textbf{99.4} & 94.3 \\
                           &                      & Base + Novel & 0.19 & 0.05 & 51.4 & 47.9 & \textbf{58.7} & \textbf{51.7} \\
\midrule

\multirow{2}{*}{MSD Pancreas} & \multirow{2}{*}{CT} & Base         & 0.07 & 0.01 & \textbf{92.5} & 62.5 & 92.1 & \textbf{62.5} \\
                              &                      & Base + Novel & 0.04 & 0.01 & 46.3 & 30.8 & \textbf{47.0} & \textbf{31.5} \\
\midrule

\multirow{2}{*}{LiTS} & \multirow{2}{*}{CT} & Base         & 2.42 & 0.61 & \textbf{98.5} & 91.9 & 98.2 & \textbf{92.8} \\
                      &                     & Base + Novel & 2.68 & 0.48 & 50.7 & 46.6 & \textbf{57.0} & \textbf{50.9} \\
                      
\midrule

\multirow{2}{*}{TotalSegmentator} & \multirow{2}{*}{CT and MRI} & Base         & 0.33 & 0.12 & 63.2 & 50.6 & \textbf{65.7} & \textbf{53.7} \\
                      &                     & Base + Novel & 0.28 & 0.16 & 43.5 & 33.9 & \textbf{46.0} & \textbf{35.5} \\
                      
\midrule
Omnis Validation Set  & All Modalities      & Base & 0.02 & 0.01 & 61.2 & 45.7 & \textbf{64.1} & \textbf{48.9} \\
\midrule

\multirow{2}{*}{Multi-Modality}   & \multirow{2}{*}{All Modalities}      & Base & 3.14 & 1.25 & 86.5 & 62.2 & \textbf{89.8} & \textbf{66.7} \\
                                  &                     & Base + Novel & 0.59 & 0.35 & 38.6 & 26.7 & \textbf{43.5} & \textbf{31.1} \\

\bottomrule
\end{tabular}
}
\caption{Comparison of zero-shot detection performance between MedROV and YOLO-World on BTCV, Cervix, MSD Liver, MSD Pancreas, LiTS, TotalSegmentator, the Omnis validation set, and multi-modality datasets, for both base and novel classes. YOLO-World + Our Omnis denotes the baseline YOLO-World model trained on our Omnis 600K dataset. Top scores are highlighted in \textbf{bold}.}
\label{tbl:table1}
\end{table*}

\section{Experiments and Results}
\label{sec:experimentsandresults}
To validate the effectiveness of MedROV, we conduct extensive experiments across a wide range of medical imaging datasets. We evaluate both detection accuracy and generalization ability under open-vocabulary and cross-modality settings, comparing MedROV against existing state-of-the-art methods and baseline models. \\
\textbf{Implementation Details:}
Starting from the baseline YOLO-World, we fine-tune MedROV on our proposed Omnis 600K dataset for 20 epochs, using 577K training and 28K validation samples. Training is performed on four NVIDIA A100 40GB GPUs with a batch size of 128, a learning rate of 0.0002, and a weight decay of 0.05. To optimize detection performance, we evaluate various combinations of IoU threshold $T$ and confidence threshold $C$ values to optimize detection performance. The best results are achieved with an IoU threshold $T$ = 0.3, and the confidence threshold of $C$ = 0.9, which effectively filters predictions to retain high-quality pseudo-labels and reduce noise.\\ 
\noindent\textbf{Evaluation Datasets and Metrics:} 
We evaluate the open-vocabulary detection (OVOD) capability of MedROV through three strategies. First, we test zero-shot transfer on entirely unseen datasets, including the Medical Segmentation Decathlon (Liver and Pancreas) \cite{antonelli2022medical} and LiTS \cite{bilic2023liver}. Second, we hold out specific classes and their images from BTCV \cite{landman2015miccai}, Cervix \cite{landman2015miccai}, TotalSegmentator CT \cite{wasserthal2023totalsegmentator} and MRI \cite{d2024totalsegmentator} to assess recognition of novel categories. Third, we evaluate cross-modality generalization using a curated Multi-Modality dataset spanning nine imaging types, created by combining Chest X-ray \cite{jaeger2014two}, Breast Lesion \cite{pawlowska2024curated}, DigestPath \cite{da2022digestpath}, PH2 \cite{mendoncca2015ph2}, CVC ClinicDB \cite{bernal2015wm}, LiverHccSeg \cite{gross2023liverhccseg}, NeurIPS CellSeg \cite{lee2023cell}, and Drishti-GS1 \cite{sivaswamy2015comprehensive}. Performance is measured using mean average precision (mAP) at IoU thresholds of 50 and 50:95. 

\begin{table*}[t]
\centering \scriptsize
\setlength{\tabcolsep}{6pt} 
\renewcommand{\arraystretch}{0.9} 
\resizebox{0.9\textwidth}{!}
{
\begin{tabular}{ l | c c c | c c | c c } 
\toprule
\rowcolor[gray]{0.85} 
\textbf{Models} $\rightarrow$  & \multicolumn{3}{c|}{\textbf{Fine-tuned}} & \multicolumn{2}{c|}{\textbf{YOLO-World + Our Omnis}} & \multicolumn{2}{c}{\textbf{MedROV (Ours)}} \\  
\rowcolor[gray]{0.85}
\textbf{Metrics} $\downarrow$  & \textbf{YOLOv8} &  \textbf{YOLOv9} & \textbf{YOLOv10} & \textbf{Zero-Shot} & \textbf{Fine-tuned} & \textbf{Zero-Shot} & \textbf{Fine-tuned} \\ 
 
\midrule
\textbf{mAP50}      & 77.3 & 74.7 & 76.2 & 74.5 & 72.2 & 79.1 & \textbf{81.9} \\  
\textbf{mAP50:95}   & 59.1 & 57.1 & 58.2 & 57.5 & 55.1 & 59.3 & \textbf{63.1} \\  
\bottomrule
\end{tabular}
}
\caption{Performance comparison of MedROV with YOLO-World + Our Omnis (baseline trained on our proposed Omnis 600K dataset) and closed-set detectors. All models are fine-tuned on the BTCV training set and evaluated on the BTCV test set. Top scores are highlighted in \textbf{bold}.}
\label{tbl:table4}
\end{table*}

\begin{table*}[t]
\centering \small
{
\begin{tabular}{l | l | c | c | c}
\toprule
\rowcolor[gray]{0.85}
\textbf{Models} $\rightarrow$ & & \textbf{BioMedParse \cite{zhao2024biomedparse}} & \textbf{MedROV (Ours)} & \\
\rowcolor[gray]{0.85} \textbf{Dataset} $\downarrow$ & \textbf{Modality} &  \textbf{mAP50}
 & \textbf{mAP50} & \textbf{Improvement} $\uparrow$  \\
\midrule
BTCV & CT & 9.09 & 79.1 & 70.01 \\
Cervix & CT & 6.49 & 33.8 & 27.31 \\
MSD Pancreas & CT & 24.61 & 47.0 & 22.39 \\
MSD Liver & CT & 18.68 & 58.7 & 40.02 \\
LiTS & CT & 12.63 & 57.0 & 44.37 \\
TotalSegmentator & CT and MRI & 0.52 & 46.0 & 45.48 \\
Multimodality & All Modalities & 12.84 & 43.5 & 30.66 \\
\bottomrule
\end{tabular}
}
\caption{Zero-shot detection performance comparison between MedROV (ours) and the foundation model BioMedParse \cite{zhao2024biomedparse} across multiple medical datasets spanning CT, MRI, and multi-modality settings. MedROV consistently outperforms BioMedParse, with improvements reported in the last column.}
\label{tbl:table6}
\end{table*}

\subsection{Results}

MedROV is the first open-vocabulary object detection (OVOD) framework tailored for medical imaging. Initial experiments showed that state-of-the-art OVOD methods for natural images—such as OV-DETR \cite{zang2022open}, OWL-ViT \cite{minderer2022simple}, and GLIP \cite{li2022grounded}—perform poorly on medical data, with near-zero accuracy across benchmarks. This highlights the significant domain gap and the need for medical-specific solutions. Given these limitations, we selected YOLO-World \cite{cheng2024yolo} as our starting point due to its strong performance. While its zero-shot inference was limited on medical data, fine-tuning it on our Omnis 600K dataset led to more meaningful results. We benchmarked MedROV against both traditional closed-set detectors and the previous state-of-the-art medical imaging detection model, BioMedParse \cite{zhao2024biomedparse}, under both open- and closed-vocabulary settings. MedSAM \cite{ma2024segment} was excluded from comparison, as it is a segmentation-only model that outputs binary masks without class labels, making it unsuitable for object detection. \\
\noindent\textbf{Comparison of MedROV with YOLO-World:}
MedROV outperforms the baseline YOLO-World \cite{cheng2024yolo}, which fails at zero-shot detection on medical datasets due to a significant domain gap between natural and medical images. To ensure a fair comparison, we evaluate against YOLO-World trained on our Omnis 600K dataset (YOLO-World + Our Omnis). As shown in Table \ref{tbl:table1}, MedROV consistently surpasses this stronger baseline across both base and novel classes. For example, it improves base + novel mAP50 from 74.5 to 79.1 on the BTCV dataset, from 43.5 to 46.0 on the TotalSegmentator CT and MRI dataset, and from 38.6 to 43.5 on the multi-modality dataset. Additionally, MedROV retains real-time performance (MedROV: 72 FPS vs. YOLO-World: 70 FPS), where real-time refers to inference speeds exceeding a threshold of 30 FPS, making it a practical solution for accurate and efficient medical object detection. \\
\noindent\textbf{Comparison of MedROV with Closed-set Detectors:}
Table \ref{tbl:table4} presents a comparison between MedROV, YOLO-World + Our Omnis (baseline trained on our Omnis 600K dataset), and traditional closed-set detectors (YOLOv8, YOLOv9, YOLOv10), fine-tuned on the BTCV training set for 20 epochs. In the zero-shot setting, MedROV achieves a strong mAP50 of 79.1, outperforming all fine-tuned closed-set models. When fine-tuned, MedROV further improves to 81.9 mAP50 and 63.1 mAP50:95, outperforming the best closed-set model (YOLOv8) by +4.6 mAP50 and +4.0 mAP50:95, demonstrating its superior generalization across both zero-shot and fine-tuned scenarios. \\
\noindent\textbf{Comparison of MedROV with BioMedParse:}
To the best of our knowledge, MedROV is the first to explore real-time OVOD in medical imaging. Given the absence of OVOD methods tailored for medical data, we compared MedROV with BioMedParse \cite{zhao2024biomedparse}, a recent multi-task foundation model for detection, segmentation, and recognition. 

BioMedParse is trained on a large corpus of image–mask–label triplets to recognize 82 predefined medical object types, but lacks generalization to unseen categories. Additionally, it does not produce confidence scores necessary for standard detection evaluation. To address this, we approximated prediction confidence by averaging the positive pixel values within each predicted mask. Its reliance on anatomy-specific preprocessing (e.g., intensity tuning per organ) further limits scalability across diverse object types. Moreover, BioMedParse struggles with multi-object images, often detecting only a subset of entities or assigning the same label to distinct structures—challenges stemming from its fixed vocabulary and training on single-object image–mask pairs. In contrast, MedROV, trained on 157 categories and incorporating BioMedCLIP, supports open-vocabulary detection, applies uniform preprocessing across all inputs, and handles multi-object images effectively. For a fair comparison, both models were evaluated on our test datasets using the same preprocessed inputs. 
\begin{figure*}[htbp]
\centering
\setlength{\tabcolsep}{1pt}
\renewcommand{\arraystretch}{1.0}

\newcommand{\fixedsizeimgtrim}[2]{%
  \includegraphics[width=4.2cm,height=4.2cm,keepaspectratio,trim=#1,clip]{#2}%
}

\begin{tabular}{>{\centering\arraybackslash}m{0.25cm} *{4}{>{\centering\arraybackslash}m{4.2cm}}}

& \textbf{LiTS} & \textbf{Breast Lesion} & \textbf{BTCV} & \textbf{MSD Liver} \\

\rotatebox{90}{\textbf{Ground Truth}} &
\fixedsizeimgtrim{0 0 0 0}{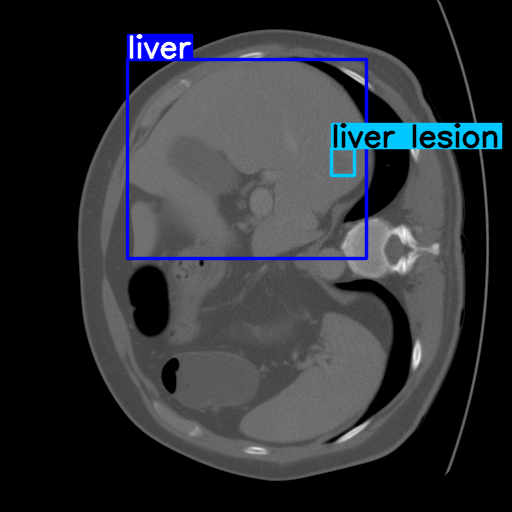} &
\fixedsizeimgtrim{0 20 0 20}{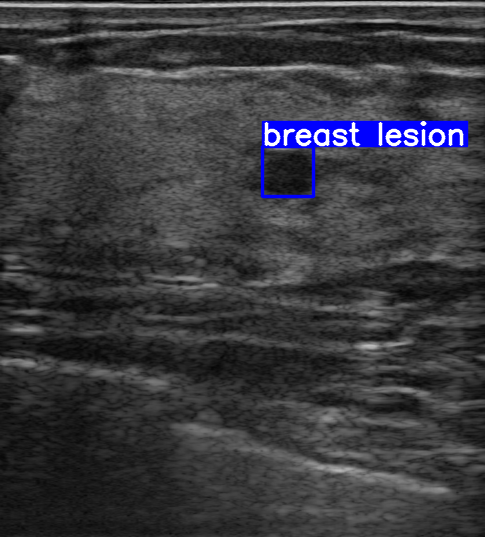} & 
\fixedsizeimgtrim{0 0 0 0}{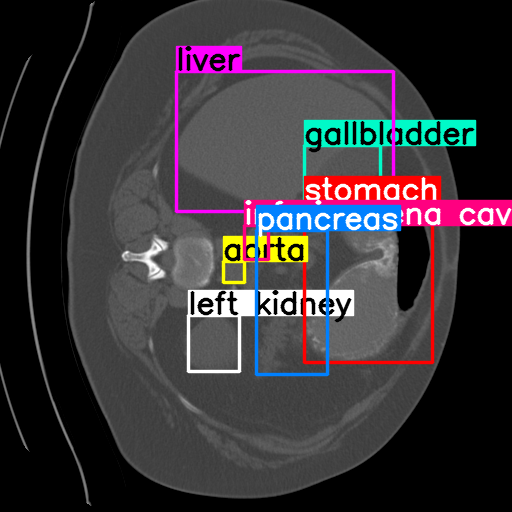} &
\fixedsizeimgtrim{0 0 0 0}{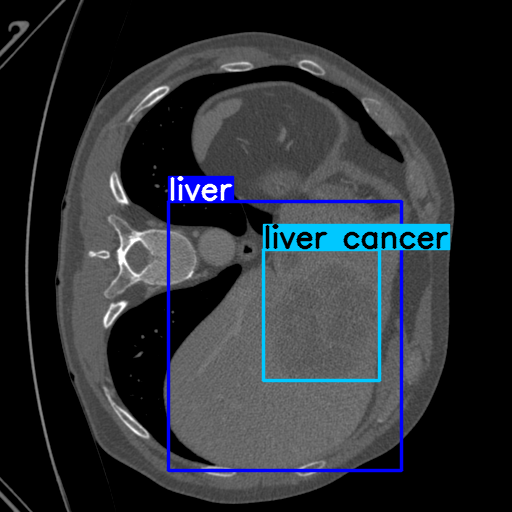} \\

\rotatebox{90}{\textbf{MedROV}} &
\fixedsizeimgtrim{0 0 0 0}{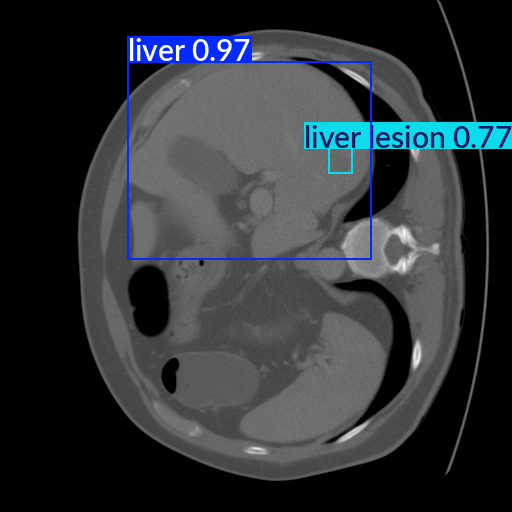} &
\fixedsizeimgtrim{0 20 0 20}{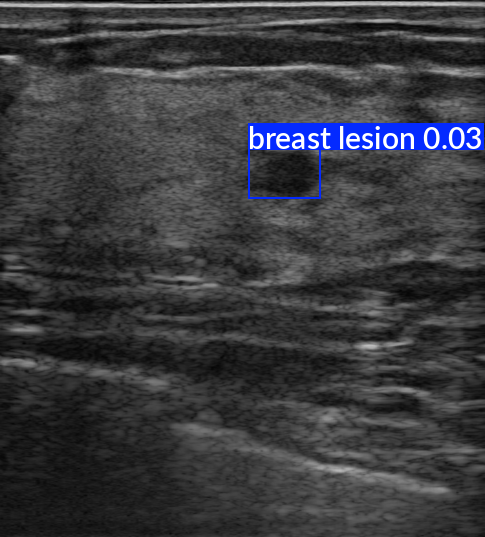} & 
\fixedsizeimgtrim{0 0 0 0}{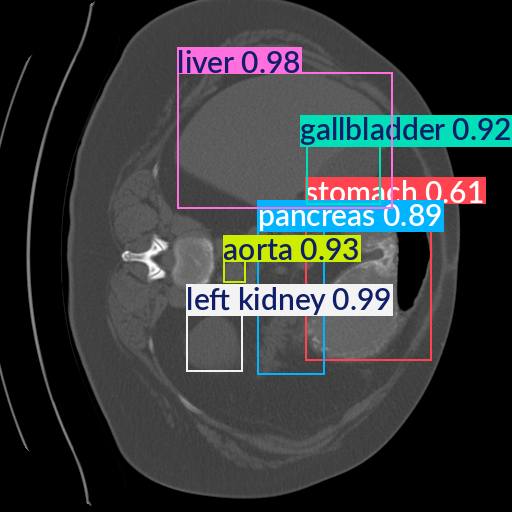} &
\fixedsizeimgtrim{0 0 0 0}{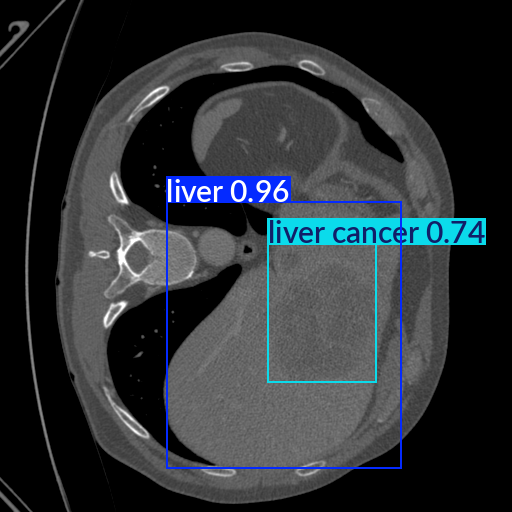} \\

\end{tabular}

\caption{Visual comparison of MedROV’s zero-shot detection performance on four datasets: LiTS, Breast Lesion, BTCV, and MSD Liver. The model successfully detects both known and novel classes, including liver lesion, breast lesion, and liver cancer.}
\label{fig:qualitative}
\end{figure*}

\begin{table*}[t]
\centering \small
{
\begin{tabular}{ l | ccc | c c | c c } 
\toprule
\rowcolor[gray]{0.85} 
 & \textbf{Omnis 600K} & \textbf{Missing} & \textbf{Enhancing} & \multicolumn{2}{c|}{\textbf{Base Classes}} & \multicolumn{2}{c}{\textbf{Base + Novel Classes}} \\ 
\rowcolor[gray]{0.85}
 & \textbf{Dataset} & \textbf{Annotations} & \textbf{Generalization} & \textbf{mAP50} & \textbf{mAP50:95} & \textbf{mAP50} & \textbf{mAP50:95} \\ 
\midrule
 Baseline        & -                   & -                     & -                             & 0.02 &  0.01  & 0.01 & 0.00 \\  
  Baseline        & \checkmark                   & -                     & -                             & 79.1 & 61.7 & 48.0 & 37.1 \\
 Baseline*        & \checkmark          & -                     & -                             & 78.8   & 61.5    & 48.4   & 37.5    \\  
 Baseline*        & \checkmark          & \checkmark            & -                             & 81.1   & 64.3   & 50.8  & 39.5  \\  
 Baseline*        & \checkmark          & \checkmark            & \checkmark                    & \textbf{81.8} & \textbf{66.7}  & \textbf{51.3} & \textbf{40.3} \\  
\bottomrule
\end{tabular}
}
\caption{Ablation study of MedROV with diverse configurations, highlighting the impact of fine-tuning the baseline on our Omnis 600K dataset, addressing missing annotations, and enhancing generalization. Results are reported for base and base + novel classes on the TotalSegmentator CT dataset. Baseline* represents MedROV with BioMedCLIP Text Encoder. Top scores are highlighted in \textbf{bold}.}
\label{tbl:table5}
\end{table*}
As shown in Table~\ref{tbl:table6}, MedROV consistently outperforms BioMedParse across all benchmarks. For example, on the BTCV dataset, it achieves an mAP50 of 79.1, compared to 9.09 by BioMedParse, an improvement of 70.01. Similar trends are observed on MSD Pancreas, LiTS, and TotalSegmentator, highlighting MedROV’s strong detection and generalization performance. Moreover, MedROV operates in real time at 70 FPS with CPU support, whereas BioMedParse runs at only 4 FPS and lacks CPU deployment capability. MedROV’s flexibility makes it better suited for real-world clinical scenarios, where broad category coverage and adaptability are essential. \\
\noindent\textbf{Qualitative Results:}
Fig.~\ref{fig:qualitative} illustrates MedROV’s zero-shot detection performance across four datasets. The model accurately identifies novel classes such as liver and breast lesions, and liver cancer, while maintaining strong performance on base classes, as seen in the BTCV dataset. Low-confidence detections (e.g., 3\% confidence for breast lesion) were retained based on observations from YOLO-World \cite{cheng2024yolo}, where novel objects can appear even with confidence scores as low as 1\%. For qualitative visualization, we use a data-driven, image-specific threshold rather than a fixed one. Specifically, we sort the detection scores in descending order and find the point where the score curve sharply bends (“elbow point”), which reflects a separation between strong and weak detections. This method is only used for visualization and does not affect quantitative evaluation, where boxes are ranked by raw confidence as in standard AP computation. Moreover, in zero-shot settings, relative confidence is often more informative than absolute values, so even low-confidence detections can be meaningful. 
\\
\noindent\textbf{Ablation study on the impact of different configurations:}
Table~\ref{tbl:table5} illustrates that fine-tuning the baseline YOLO-World on our Omnis-600K dataset significantly improves performance compared to the zero-shot baseline. Replacing the CLIP text encoder with BioMedCLIP (Baseline*) results in a slight decrease in base class performance, with mAP50 dropping from 79.1 to 78.8, but improves novel class performance from 48.0 to 48.4. This reflects better alignment with medical terminology, as BioMedCLIP is trained on a large dataset of medical image–text pairs. Despite the minor tradeoff, the BioMedCLIP text encoder is used in all experiments due to its domain relevance. Adding pseudo-labeling to address missing annotations further improves the Base + Novel mAP50 from 48.4 to 50.8. Finally, incorporating BioMedCLIP image features to enhance generalization leads to the highest performance, achieving 81.8 mAP50 on base classes and 51.3 mAP50 on base + novel classes.
\section{Conclusion}
\label{sec:conclusion}
We introduce MedROV, the first real-time open-vocabulary detection method for medical imaging. By adapting YOLO-World to the medical domain, we replace the CLIP text encoder with BioMedCLIP and curate Omnis, a large-scale dataset comprising 600K samples across nine imaging modalities. To address missing annotations when merging datasets, we employ a pseudo-labeling strategy. We also integrate knowledge from a foundation model to enhance generalization. MedROV outperforms the baseline YOLO-World, recent foundation models, and closed-set detectors in open-vocabulary performance, while maintaining real-time speed. Future work will focus on curating an open-vocabulary test set with a broader range of categories and extending the approach to 2D and 3D segmentation tasks.

\subsubsection*{Acknowledgments}
\label{sec:acknowledgment}
This work is partially supported by the MBZUAI-WIS Joint Program for AI Research (Grant number WIS-P008) and the NVIDIA Academic Grant 2025.

{
    \small
    \bibliographystyle{ieeenat_fullname}
    \bibliography{main}
}

\end{document}


\maketitle
\begin{figure*}[!b]
  \centering
  \captionsetup[sub]{justification=centering}
  \begin{subfigure}[t]{0.31\textwidth}
    \centering
    \includegraphics[height=0.22\textheight, keepaspectratio]{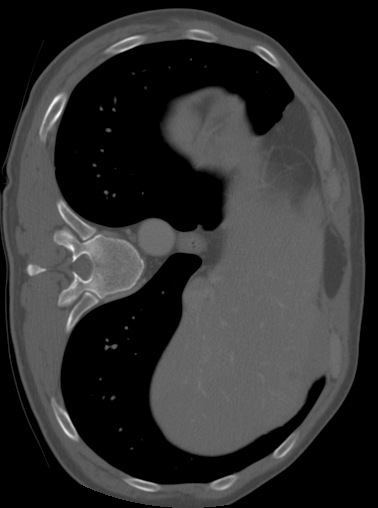}
    \caption{No detections}
    \label{fig:fail_no_dets}
  \end{subfigure}\hfill
  \begin{subfigure}[t]{0.31\textwidth}
    \centering
    \includegraphics[height=0.22\textheight, keepaspectratio]{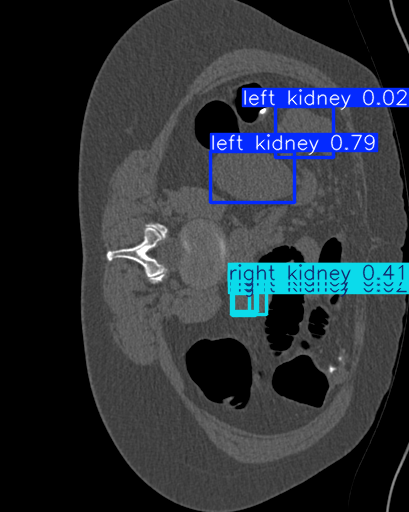}
    \caption{Multiple/Duplicate detections}
    \label{fig:fail_multi}
  \end{subfigure}\hfill
  \begin{subfigure}[t]{0.31\textwidth}
    \centering
    \includegraphics[height=0.22\textheight, keepaspectratio]{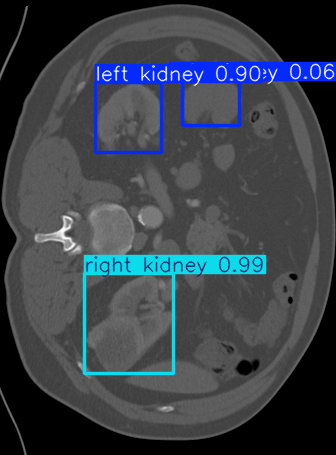}
    \caption{Multiple/Duplicate detections}
    \label{fig:fail_lr1}
  \end{subfigure}

  \vspace{6pt}

  \begin{subfigure}[t]{0.31\textwidth}
    \centering
    \includegraphics[height=0.22\textheight, keepaspectratio]{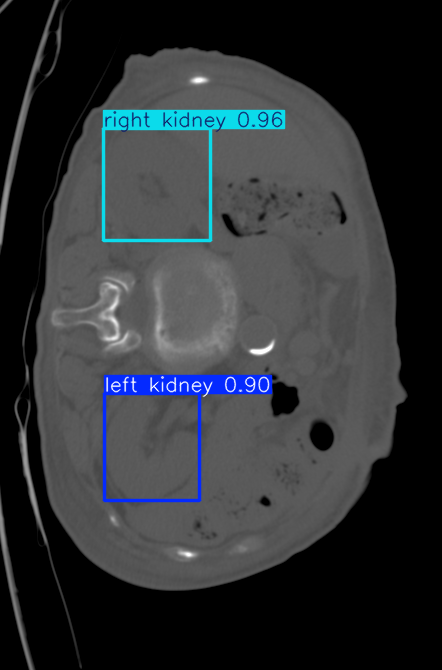}
    \caption{Right/Left Kidney Confusion}
    \label{fig:fail_conf_multi}
  \end{subfigure}\hspace{0.08\textwidth}
  \begin{subfigure}[t]{0.31\textwidth}
    \centering
    \includegraphics[height=0.22\textheight, keepaspectratio]{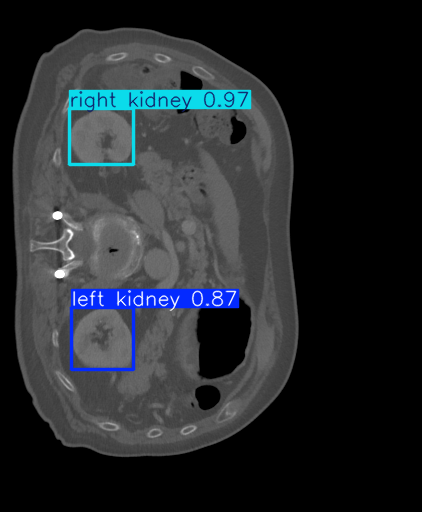}
    \caption{Right/Left Kidney Confusion}
    \label{fig:fail_lr2}
  \end{subfigure}

  \caption{\textbf{Failure cases analysis.} Typical errors include: missed detections (a),  duplicate boxes (b and c), left/right kidney confusion (d and e).}
  \label{fig:failure_grid}
\end{figure*}

\noindent\textbf{Analysis of Failure Cases:}
Figure~\ref{fig:failure_grid} illustrates failure modes observed in MedROV’s predictions. Panel (a) shows a missed detection, where the model fails to identify visible organs such as the lungs, liver, vertebrae and ribs. Panels (b) and (c) depict duplicate detections, with multiple overlapping boxes assigned to the same left/right kidney. Panels (d) and (e) highlight left–right kidney confusion, where the model swaps the anatomical labels of the kidneys. These errors may arise from anatomical symmetry or ambiguous visual cues during training. Addressing such cases will be our future work.

\vspace{-2cm}

